%% file: main.tex
\documentclass[10pt,twocolumn,letterpaper]{article}

\usepackage{3dv}
\usepackage{times}
\usepackage{epsfig}
\usepackage{graphicx}
\usepackage{amsmath}
\usepackage{amssymb}
\usepackage{algorithm}
\usepackage{amsthm}
\usepackage{booktabs}
\usepackage{multirow}
\usepackage{mathrsfs}
\usepackage{algpseudocode}

\usepackage[hypcap]{caption}
\usepackage{subcaption}
\usepackage{stackengine}
\usepackage{array}
\usepackage{bigstrut}
\usepackage{boldline}

\usepackage[pagebackref=true,breaklinks=true,colorlinks,bookmarks=false]{hyperref}
\threedvfinalcopy %

\def\threedvPaperID{33} %

\ifthreedvfinal\fi

\begin{document}

\title{GAMesh: Guided and Augmented Meshing for Deep Point Networks}

\author{
Nitin Agarwal \hspace{10em} M Gopi \\
University of California, Irvine\\
}

\newcommand{\STAB}[1]{\begin{tabular}{@{}c@{}}#1\end{tabular}}
\newcommand\T{\rule{0pt}{1ex}}                                          %
\newcommand\B{\rule[-1.5ex]{0pt}{0pt}}                                  %
\renewcommand{\labelitemi}{\textbullet}
\newcommand{\rulesep}{\unskip\ \vrule\  }

\maketitle

\input{tex_files/abstract.tex}

\input{tex_files/intro.tex}

\input{tex_files/related_work.tex}

\input{tex_files/method.tex}

\input{tex_files/experiments.tex}

\input{tex_files/conclusion.tex}

\newpage
{\small
\bibliographystyle{ieee}
\bibliography{refs}
}

\end{document}

%% file: tex_files/abstract.tex
\begin{abstract}
We present a new meshing algorithm called guided and augmented meshing, GAMesh, which uses a mesh prior to generate a surface for the output points of a point network. By projecting the output points onto this prior and simplifying the resulting mesh, GAMesh  ensures a surface with the same topology as the mesh prior but whose geometric fidelity is controlled by the point network. This makes GAMesh independent of both the density and distribution of the output points, a common artifact in traditional surface reconstruction algorithms. We show that such a separation of geometry from topology can have several advantages especially in single-view shape prediction, fair evaluation of point networks and reconstructing surfaces for networks which output sparse point clouds. We further show that by training point networks with GAMesh, we can directly optimize the vertex positions to generate adaptive meshes with arbitrary topologies. Code and data are available on the project webpage\footnote{\url{https://www.ics.uci.edu/~agarwal/GAMesh}}.
\end{abstract}

%% file: tex_files/intro.tex
\section{Introduction}
\label{intro}

Meshes are a natural choice for representing 3D shapes as they can describe complex topologies and surface details while being memory efficient. They are also essential to applications like rendering, simulation, shape analysis and 3D printing to name a few. In this work, we focus on reconstructing surface meshes for point sets which are the output of point reconstruction networks. 

Generating surfaces from point clouds has a rich history in computer graphics. Loosely speaking, surface reconstruction requires computing a surface for a set of input points such that those points lie on the reconstructed surface \cite{Berger:2013aa,Berger:2017aa,Gopi:2000aa}. Such points usually come from 3D range scan data which inherently have noise \cite{Berger:2017aa}. For this reason most reconstruction algorithms tend to generate an approximate surface where the input points are not the vertices of the final mesh \cite{Kazhdan:2006aa,Kazhdan:2013aa,Alexa:2003aa,Ohtake:2003aa}.

However, with the recent success of deep point networks \cite{Qi:2017aa,Qi:2017ab}, many learning based geometric algorithms have started to output 3D point clouds \cite{Fan:2017aa,Achlioptas:2017aa,Lin:2018ab}. Reconstructing a surface connecting all these output points would not only help in visualizing and evaluating these point networks but could also assist in training them end-to-end for 3D surface prediction.

Keeping these ideas in mind, we propose \emph{GAMesh - guided and augmented meshing}, an automatic meshing algorithm for point clouds (``reconstructed/output points") which are typically the output of point based networks. GAMesh guarantees correct geometry and topology of the final surface while preserving as much geometric features as maintained by the output points of the network. Furthermore, GAMesh can be used both in post-processing to mesh the output points or to train the point network to directly optimize the vertex positions of the final 3D mesh.

\begin{figure}[!t]
\centering
\captionsetup[sub]{font=scriptsize,justification=raggedright}
\includegraphics[trim={1cm 3cm 1cm 1cm},clip,width=\linewidth]{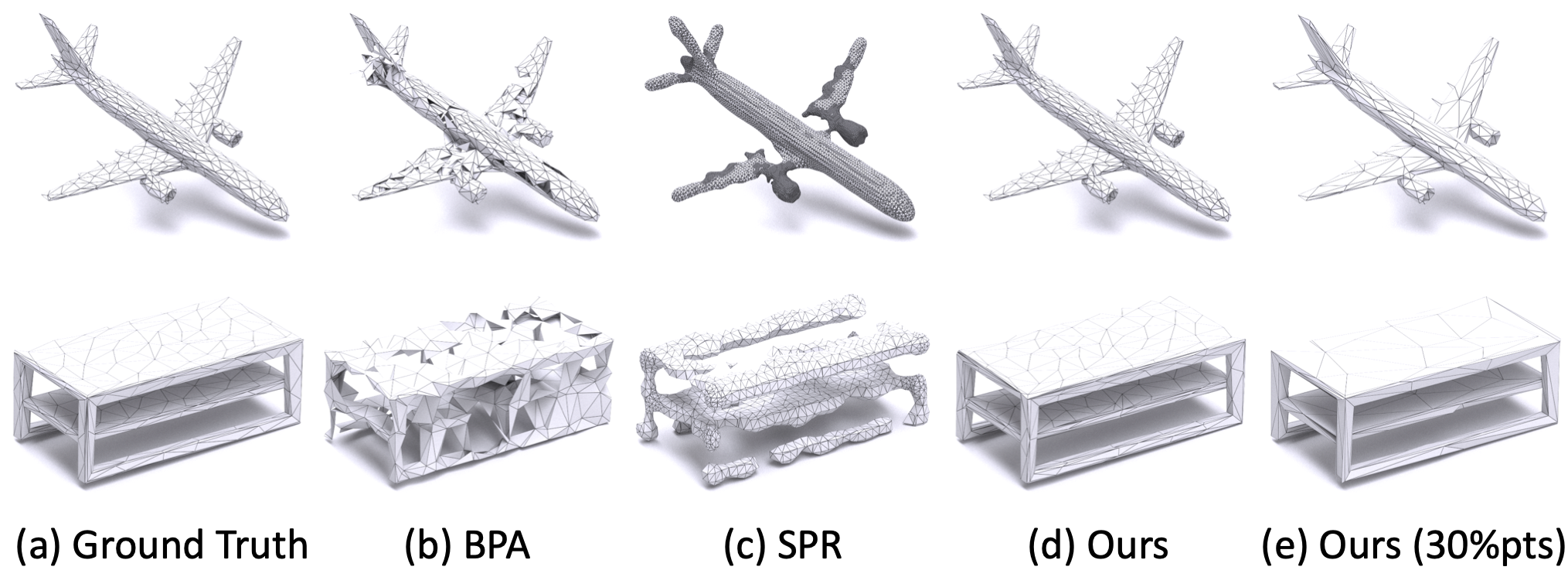}
\begin{subfigure}[!t]{0.21\linewidth}
    \caption{Ground Truth}
\end{subfigure}
\begin{subfigure}[!t]{0.18\linewidth}
    \caption{BPA}
\end{subfigure}
\begin{subfigure}[!t]{0.17\linewidth}
    \caption{SPR}
\end{subfigure}
\begin{subfigure}[!t]{0.17\linewidth}
    \caption{Ours}
\end{subfigure}
\begin{subfigure}[!t]{0.22\linewidth}
    \caption{Ours(30\%pts)}
\end{subfigure}
\caption{\textbf{Surface Reconstruction.} Surfaces reconstructed using all the original vertices with (b) BPA (c) SPR and (d) our method. We reconstruct the same mesh as the input mesh and preserve the geometry and topology even with 30$\%$ of the original points. Please zoom in for details.}
\label{fig:GAM_vs_bpa/spr}
\end{figure}

The key insight of GAMesh is to use a mesh prior with similar topology as the ground truth shape for reconstructing a surface for the output points. In our method, the output mesh is \emph{guided} by the mesh prior through a process of \emph{augmenting} the mesh prior vertices with the output points, and iteratively removing all the mesh prior vertices while retaining only its topology. This process results in an output mesh that has only the output points but with the topology of the mesh prior. Hence, unlike traditional surface reconstruction problems where the objective is to reconstruct an approximating surface using \emph{only} a point cloud, our goal is to use both a mesh prior and the output points to generate a surface which faithfully preserves the topology and the output geometric features. We demonstrate this through single-view 3D reconstruction (SVR) where we use meshes from implicit networks as mesh priors, to generate surfaces for the output points of a point reconstruction network. By combining the outputs of these two networks, we show that the result not only performs better than the individual reconstruction methods, but also outperforms priors works (Fig. \ref{fig:svr_test}). Further, while most SVR methods fail to exploit the mesh representation and generate surfaces with uniform distribution of vertices, even in low curvature and planar regions, we show that by training point reconstruction networks with GAMesh, we can optimize the vertex position to generate adaptive meshes \cite{Smith:2019aa} with arbitrary topology. 

In addition to SVR, we show the advantages of GAMesh for evaluating point networks. Unlike existing surface reconstruction methods like Ball-Pivot Algorithm (BPA) \cite{Bernardini:1999aa} and Screened-Poisson Reconstruction (SPR) \cite{Kazhdan:2013aa}, GAMesh is insensitive to both density and distribution of points. This makes GAMesh an ideal candidate for meshing the output points from a variety of point networks \cite{Wang:2020ab,Loizou:2020aa,Agarwal:2019aa}. Using ground truth shapes as mesh priors, we show that surfaces reconstructed by GAMesh are more reliable and should be used to analyze the performance of point reconstruction networks rather than solely relying on the output points.

%% file: tex_files/related_work.tex
\section{Related Works}

\noindent
\textbf{Surface Generation for Deep Networks.} One straightforward way of reconstructing a surface for the output points from a deep network is for the network to implicitly encode the point ordering such that the same input connectivity can be used for the output points. Such networks are typically mesh based networks \cite{Litany:2017aa,Tan:2018ac,Ranjan:2018aa,Tan:2018ab} which use L1 or L2 reconstruction loss between the input and output points. These methods require the number of points in the input and output to be the same. Other methods use a fixed template mesh which is deformed to match the ground truth 3D shape and later the connectivity of the template is used as a surrogate for the output points \cite{Groueix:2018ab,Wang:2018aa,Kanazawa:2018aa,Gao:2019aa,Smith:2019aa}. Although these methods achieve impressive results, they limit the topology of the reconstructed mesh to that of the template. Further, extending these approaches to existing point networks by enforcing one-to-one correspondence (i.e. via L1 or L2 loss) reduces the network's performance.

Point based networks, which do not enforce one-to-one correspondences between the input and output points, use analytical surface reconstruction techniques like BPA and SPR to reconstruct surfaces \cite{Yu:2018ac,Yu:2018ab,Groueix:2018ab,Rakotosaona:2019aa}. While BPA does not introduce new points, making it an ideal choice for surface generation, it cannot guarantee correct topology. It is extremely sensitive to the point distribution and radius of the pivoting ball and can even produce results without connecting all the points (Fig. \ref{fig:GAM_vs_bpa/spr}b). On the other hand, SPR is an interpolating technique which is robust to noise but cannot guarantee correct geometry as it fails to capture sharp features (Fig. \ref{fig:GAM_vs_bpa/spr}c). Further, SPR requires accurate normals, which if asked from the network, diminishes the network's performance \cite{Groueix:2018ab}. Although other reconstruction \cite{Calakli:2011aa,Gopi:2000aa,Ohtake:2003aa} and remeshing \cite{Botsch:2004aa} techniques have been proposed, they are more complex and non-trivial to be incorporated inside point networks.

We propose a simple surface generation algorithm that can be used both to post-process the output points of point networks and can be incorporated inside these point network to train them end-to-end for surface prediction. Our method guarantees correct geometry and topology by connecting all the output points and does not require any additional information such as normals. It only uses a mesh prior as a guide which depending on the application, can either be generated using other reconstruction methods or is already present. Furthermore, our method does not introduce any additional points and is insensitive to both the number and distribution of output points. It even works on non-manifold meshes, a common feature in popular 3D shape repositories like ShapeNET \cite{shapenet2015}.

\noindent
\textbf{Single View Reconstruction.} A variety of representations have been explored for predicting 3D shapes from a single view image.

\emph{Voxels \& Points:}  Methods which reconstruct voxels \cite{Choy:2016aa,Wu:2016aa,Yan:2016aa}, while intuitive, are often limited by voxel resolution, resulting in missing details. Although techniques like Octree \cite{Riegler:2017aa,Wang:2017aa} help scale to higher resolutions, they are complex and usually suffer from voxel-based discretization artifacts. Points are popular alternatives as they are light weight, flexible and can describe fine geometry \cite{Fan:2017aa,Lin:2018ab,Yang:2018aa}. However, they require meshing \cite{Bernardini:1999aa,Kazhdan:2013aa} as a post-processing step to generate the actual 3D surface. Further, these meshing methods often require heavy parameter tuning to reconstruct surfaces with correct topology and geometry. 

\emph{Fixed Topology Methods:} A few methods \cite{Wang:2018aa,Smith:2019aa} use graph convolutions to directly predict triangle meshes. They use template models for 3D supervision which constraints the topology of the final mesh to that of the template (usually genus zero). Impressive 3D shape reconstruction has also been realized without 3D supervision \cite{Chen:2019ag, Liu:2019ad, Kato:2018aa,Kanazawa:2018aa,Kar:2015aa}. However, they still rely on fixed or category-specific templates restricting the final mesh topology.

\begin{figure*}[!t] 
\centering
\includegraphics[trim={0cm 0cm 0cm 0cm},clip,width=0.97\textwidth]{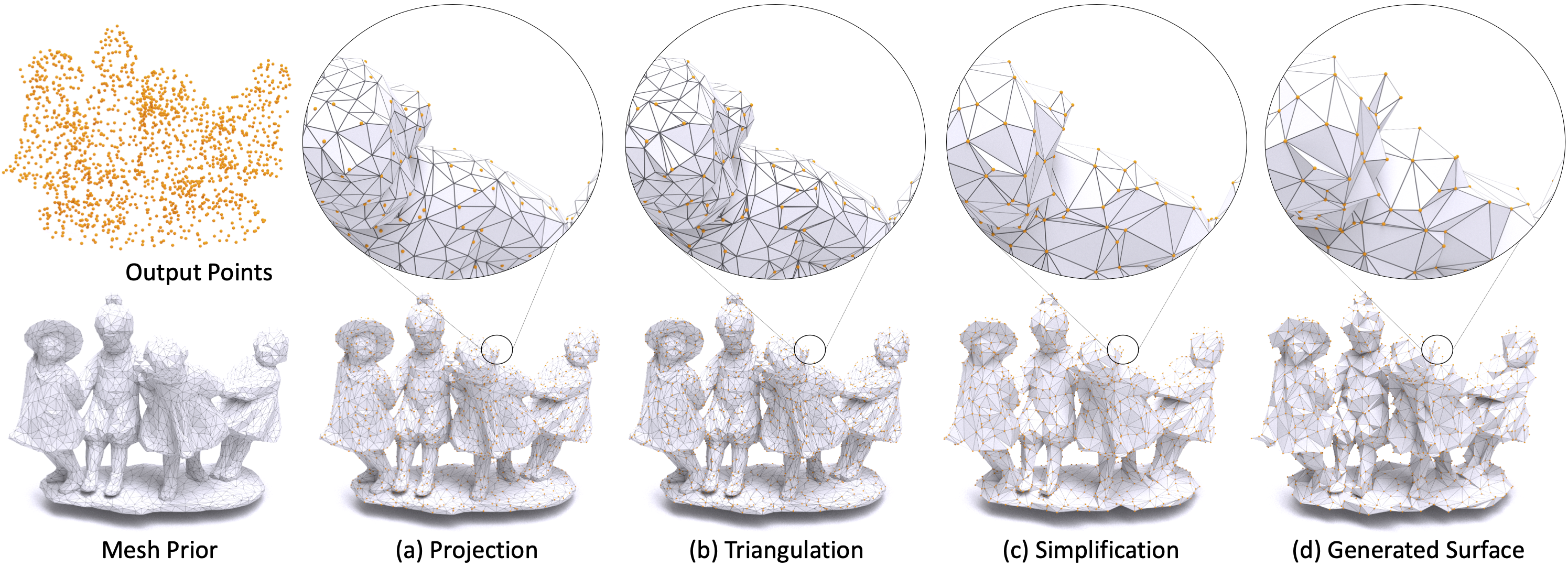}
\caption{\textbf{Surface reconstruction using GAMesh.} Our method uses both the output points (orange) from a point network and a mesh prior to generate a surface for the output points. For the sake of illustration, here we use the ground truth mesh as priors for GAMesh. (a) Projection of output points on the mesh surface is followed by (b) Triangulation to ensure that they are incorporated into the mesh prior. (c) Simplification via sequential edge collapse removes all the original points. (d) Unprojection of the points recovers the final surface for the output point set. Please zoom in for details.}
\label{fig:overview}
\end{figure*}

\emph{Arbitrary Topology Methods:} There have been efforts to generate meshes with complex topologies. Pan et al. \cite{Pan:2019aa} propose a topology modification network which progressively removes faces that have high error from a zero genus template. Methods which represent shapes using multiple, possibly overlapping patches \cite{Groueix:2018ab,Wang:2018ae} can also reconstruct models with arbitrary topologies. However, these intersecting, overlapping patches with open boundaries are not suitable for downstream applications. Further, there are works which formulate SVR as a two-stage problem - shape retrieval and deformation \cite{Pontes:2017aa,Kurenkov:2018aa,Kong:2017aa}. While such methods can also produce meshes with arbitrary topology, they are limited to the diversity of shapes they can reconstruct by the models in the repository.

\emph{Implicit Methods:} Recently, signed distance functions \cite{Park:2019aa,Xu:2019aa,Chen:2018aa} have also been used for 3D shapes. Although these methods can represent meshes with arbitrary topology, they fail to capture fine details and reconstruct smooth meshes. They optimize auxiliary losses defined on intermediate representations and require an additional post-processing step like marching cubes \cite{Lorensen:1987aa} for mesh extraction. Further, they generate dense uniform points even in low curvature and planar regions and thus do not exploit the advantages of a mesh representation.

\emph{Hybrid Methods:} Hybrid methods combine the benefits of two or more representations. Liao et al. \cite{Liao:2018aa} combine voxel and mesh representation by proposing a differentiable marching cubes algorithm to convert the output of a volume decoder to a mesh and optimized the network using geometric losses. Gkioxari et al. \cite{Gkioxari:2019aa} improved upon this by first obtaining a coarse voxel prediction with correct topology and then refining it using graph convolutions to recover details. Tang et al. \cite{Tang:2019aa} proposed to combine all three - point, voxel and mesh representations for SVR where the skeletal points predicted from the RGB image is converted to a coarse volume using voxelization before being refined using a series of 3D and graph convolutions to produce an output mesh.

\emph{Our Approach:} We propose a surface reconstruction technique that combines the benefits of point and implicit representation. We combine the output of point networks which are good at geometry (does not have topology), and meshes from implicit networks which are good at topology (maybe bad at geometry) to reconstruct meshes with both high fidelity and correct topology. Hence, unlike previous hybrid techniques which first obtain a coarse mesh from a voxel predictor and later refine that mesh to get details, GAMesh disentangles geometry from topology, by making the point network solely responsible for geometry and the mesh prior responsible for topology. Further, by training point networks with GAMesh, we optimize the vertex positions to generate adaptive meshes with arbitrary topology.

%% file: tex_files/method.tex
\section{Guided and Augmented Meshing}

Our goal is to compute a surface for the output points from a point network using a mesh prior which coarsely resembles the ground truth shape and has correct topology. Depending on the application, such mesh priors can either be obtained using other reconstruction methods (e.g implicit methods) or are already present (e.g ground truth meshes). We propose to use these mesh priors to guide the reconstruction of an output mesh with the output points. Such a reconstruction is more accurate in terms of topology and geometry than an off-the-shelf surface reconstruction algorithm like BPA or SPR. If the output points from a point network are exactly the same as the vertices of the ground truth mesh, we want the reconstructed mesh also to be exactly the same as the ground truth mesh. Fig. \ref{fig:GAM_vs_bpa/spr}d shows such a case where using the input vertices we reconstruct the ground truth mesh, whereas other methods do not. For any other output point sets, we would like the reconstructed mesh to best preserve the input geometric features and topology. To this end, we propose a new meshing algorithm called \emph{guided and augmented meshing}, GAMesh, for which the input is a mesh prior and the reconstructed points from a point network and the output is a triangular mesh connecting those reconstructed points. GAMesh works with both manifold and non-manifold mesh priors, and with both dense and sparse reconstructed points. Furthermore, it does not require the reconstructed points to have the same cardinality as the mesh prior. These properties make GAMesh suitable for a wide variety of 3D deep learning point networks.

GAMesh has two main algorithmic steps. First, the mesh prior is \emph{augmented} by projecting the reconstructed points onto the mesh prior and retriangulating to include these projected points in the mesh prior. This ensures that the topology of the mesh prior is carried over to GAMesh's output. Second, all the original points in the augmented mesh are removed by collapsing them to the nearest projected points, resulting in a mesh with only the projected points as vertices and with the connectivity that is \emph{guided} by the mesh prior. Finally, the projected points are ``unprojected", yet retaining their mesh connectivity to get the final meshing of the reconstructed points as shown in Fig. \ref{fig:overview}. We now discuss these two steps in detail.

\begin{figure}[!t]
\centering
\includegraphics[width=0.95\linewidth,height=3.2cm]{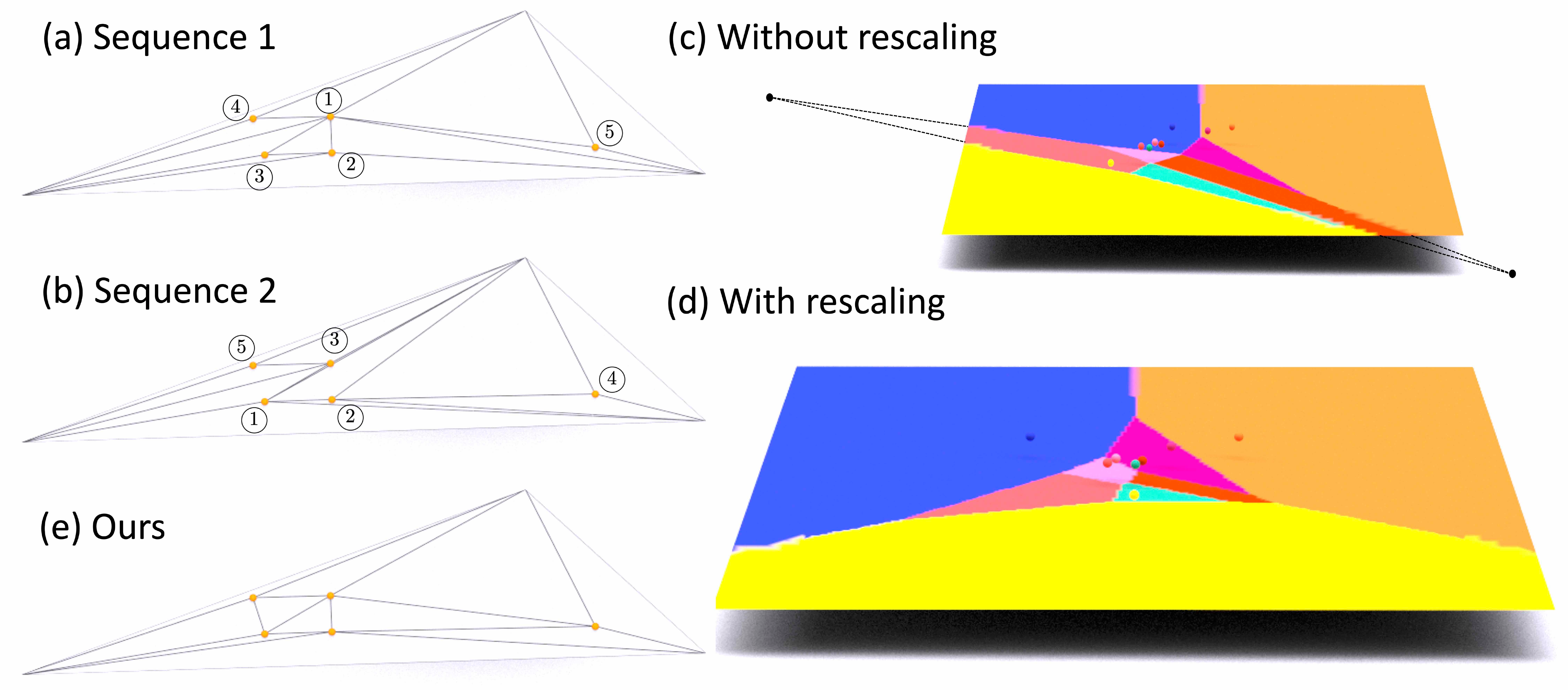}
\caption{\textbf{Projection.} (a-b) Sequential processing of points leads to two different triangulations for the same set of projected points. (c) For a triangle, we map all the projected points and its base vertices to a dense 2D grid. To prevent Voronoi vertices (black points) from lying outside the grid, (d) we rescale the points to an equilateral triangle. (e) We then compute the Voronoi diagram and use it to generate Delaunay triangulation for the projected points. Please zoom in for details.}
\label{fig:projection}
\vspace{-3mm}
\end{figure}

\subsection{Projection}
GAMesh induces the topology of the mesh prior into the output mesh by first projecting the reconstructed points onto the mesh prior and retriangulating the mesh. Hence, unlike other surface reconstruction algorithms that starts with no connectivity, GAMesh starts with the connectivity of the mesh prior and finds the final output connectivity by removing unwanted edges.

Once the points are projected onto the mesh prior, they have to be connected to existing points in the mesh to create a valid triangulation. A valid triangulation could be achieved through sequential projection of each point. However, such a processing is not only slow, but leads to inconsistent and bad quality triangulation as shown in Fig. \ref{fig:projection}(a-b). We propose an order-independent Delaunay triangulation based on 2D grid where we process all the projected points inside a triangle in parallel. Our method is fast, differentiable and can easily be incorporated inside point networks (Sec. \ref{svr_online}) unlike other triangulation techniques \cite{Barber:1996aa,Shewchuk:2002aa}.

Given the vertices of a base triangle and the projected points which lie inside the triangle, we first map all these points to a dense 2D grid. We then compute for each grid point the closest mapped point. We observe that a small neighborhood of grid points that is closest to three different mapped points contains a Voronoi vertex, which in turn indicates a Delaunay triangle connecting the corresponding three closest mapped points. To prevent Voronoi vertices from lying outside the 2D grid, we rescale the base triangle to an equilateral triangle and map all the projected points to their corresponding barycentric coordinates before computing the above mapping as shown in Figure \ref{fig:projection}(c-d). In order to ensure that the triangulation is computed fast, we perform the following optimizations - a) All points that project on the edges of the base triangle are perturbed slightly to fall inside either of the neighbouring triangles. This allows us to process all the triangles independently and in parallel. b) We only process triangles which contain at least one projected point. c) Triangles with only one projected point are directly split into 3 triangles. d) Triangles with more than one projected point are processed using the above projection-to-grid method. 

In all cases, the projection operation refers to finding the closest point on the mesh prior, and moving the reconstructed point to that closest point on the mesh. This definition of projection gracefully handles projecting points on a mesh with boundaries as well. Finally, if the input mesh is a non-manifold mesh, all the above operations and definitions work without any modification.

\begin{table*}[!t]
\centering
\caption{\textbf{GAMesh vs Points/Implicit Networks.} F1-scores on ShapeNet testset where we reconstruct meshes using GAMesh which combines geometry from a point network (PSG$^{+}$) and topology from implicit networks (OccNET or IM-NET). Such a combination performs better (shown in bold) than either the implicit network (columns 2 \& 4) or the point network. Using the ground truth meshes as prior for GAMesh gives an upper bound for the same output points.}
\resizebox{\linewidth}{!}{
  \begin{tabular}{l |c|cc|cc|c|c|cc|cc|c}
    \toprule
      & \multicolumn{6}{c|}{F1$^{\tau}$} & \multicolumn{6}{c}{F1$^{2\tau}$} \\
     \cmidrule(lr){2-7} \cmidrule(lr){8-13}
     \multirow{2}{*}{Category} & \multirow{2}{*}{PSG$^{+}$} & \multicolumn{2}{c|}{OccNET} & \multicolumn{2}{c|}{IM-NET} & GT Mesh & \multirow{2}{*}{PSG$^{+}$} & \multicolumn{2}{c|}{OccNET} & \multicolumn{2}{c|}{IM-NET} & GT Mesh\\
    \cmidrule(lr){3-4} \cmidrule(lr){5-6} \cmidrule(lr){9-10} \cmidrule(lr){11-12}
     &  & \cite{Mescheder:2019aa}  & w/ GAMesh & \cite{Chen:2018aa} & w/ GAMesh & w/ GAMesh &   & \cite{Mescheder:2019aa} & w/ GAMesh & \cite{Chen:2018aa} & w/ GAMesh & w/ GAMesh  \\
    \midrule
    Plane & 82.83 & 72.24 & \textbf{85.46} & 81.90 & \textbf{85.76} & 88.01       & 91.70 & 82.27 & \textbf{91.99} & 89.41 & \textbf{92.13} & 93.68\\
    Bench & 64.07 & 67.67 & \textbf{73.03} & \textbf{75.50} & 74.07 & 76.45     & 81.86 & 80.93 & \textbf{83.66} & \textbf{86.41} & 84.55 & 86.06\\
    Cabinet & 44.83 & 67.54 & \textbf{71.06} & 67.54 & \textbf{70.90} & 72.99           & 70.91 & 79.90 & \textbf{84.17} & 80.88 & \textbf{84.26} & 86.37\\
    Car & 55.48 & 59.70 & \textbf{76.02} & 63.48 & \textbf{76.41} & 80.45               & 80.40 & 73.54 & \textbf{88.23} & 76.36 & \textbf{88.51} & 91.25\\
    Chair & 51.61 & 64.14 & \textbf{68.29} & 67.40 & \textbf{69.02} & 71.98            & 74.45 & 77.61 & \textbf{81.36} & 80.93 & \textbf{82.04} & 84.32\\
    Monitor & 45.16 & 58.89 & \textbf{64.06} & 60.75 & \textbf{63.03} &  67.49         & 70.11 & 73.11 & \textbf{78.56} & 74.72 & \textbf{78.26} & 81.04\\
    Lamp  & 48.10 & 47.84 & \textbf{56.34} & 54.15 & \textbf{57.19} &  60.79             & 66.02 & 58.24 & \textbf{68.45} & 66.03 & \textbf{69.20} & 72.45 \\
    Speaker  & 36.67 & 50.86 & \textbf{57.69} & 56.33 & \textbf{59.60} &  61.50          & 61.79 & 63.80 & \textbf{73.01} & 70.41 & \textbf{74.37} & 76.04\\
    Firearm  & 81.69 & 72.95 & \textbf{82.35} & 78.36 & \textbf{83.02} & 85.70          & 91.39 & 84.36 & \textbf{89.74} & 88.51 & \textbf{90.28} & 92.30\\
    Couch & 42.81 & 60.77 & \textbf{64.98} & 63.34 & \textbf{66.16} & 67.70             & 68.50 & 74.08 & \textbf{79.38} & 76.38 & \textbf{80.10} & 81.14\\
    Table & 60.92 & \textbf{71.65} & 70.09 & \textbf{72.12} & 69.60 & 72.08             & 80.26 & \textbf{81.26} & 82.18 & \textbf{83.29} & 81.84 & 84.68\\
    Cellphone  & 63.98 & 76.78 & \textbf{79.03} & 78.16 & \textbf{80.26} & 81.34         & 84.49 & 86.62 & \textbf{89.38} & 87.23 & \textbf{89.96} & 89.97 \\
    Watercraft & 55.13 & 46.84 & \textbf{66.09} & 60.93 & \textbf{65.54} & 70.14          & 76.09 & 60.70 & \textbf{79.75} & 74.82 & \textbf{79.34} & 83.17 \\
    \midrule
    Mean & 56.41 & 62.91 & \textbf{70.39} & 67.68 & \textbf{70.82} & 73.59             & 76.77 & 75.11 & \textbf{82.30} & 79.65 & \textbf{82.68} & 84.80\\
    \bottomrule
  \end{tabular}
}
\label{tab:gam_performance}
\end{table*}

\subsection{Simplification}
Using the mesh generated from the previous step, we next perform simplification where we remove all the original vertices (from mesh prior) through edge collapse operations so that the final mesh contains only the projected vertices. As the order of the edge-collapses determine the final connectivity of the mesh, we sequence the edge collapses based on a cost function.

In order to simplify the mesh we first label all the projected points as 0 and the original points as 1 and select all the edges between the original points and the original and the projected points. Any edge between two projected points is not collapsed. The final point after the collapse of an edge between (a) two original points is, for the sake of simplicity, set to be the mid point of that edge and that mid point is again labelled as original point, and (b) an original and a projected point is set to be the projected point since we cannot move the projected point. Based on these final positions of the points the cost for edge collapse operations is computed as:

\begin{equation}
\Delta(v) = e^{(l_1 + l_2)}||v_1 - v_2||_2^2
\label{equ:cost}
\end{equation}
\noindent
where $v_1$ and $v_2$ are the vertices forming an edge and $l_1$ and $l_2$ are their corresponding labels. Edges are placed in a priority heap with minimum cost edge on top, and are iteratively removed from the top and collapsed, until the heap is empty. After each edge collapse, the cost to collapse the new edges are computed and they are added to the heap if they are not between two projected points.

Although this works well in practice, there could be few edge collapses that cause triangle flips. In such cases, that edge collapse is skipped, edge removed from the heap, and revisited again after other edge collapses. Note, that an edge collapse that introduces triangle flips may become a valid edge collapse after other neighborhood edge collapses. However, at the end, all original vertices are removed through edge collapses irrespective of whether it creates flipped triangles or not (Fig. \ref{fig:overview}c). Although quadric error \cite{Garland:1997aa} could be used as a cost function, we empirically found our weighted edge length cost (Eq. \ref{equ:cost}), which favors short edges, reduces the number of triangle flips. Finally, the projected points are unprojected back to their original reconstructed point positions to get a final mesh (Fig. \ref{fig:overview}d).

%% file: tex_files/experiments.tex
\section{Experiments}

We analyze the effectiveness of GAMesh for single-view 3D reconstruction in two settings. First, we use GAMesh in post-processing to combine the outputs of a point network and an implicit network and compare the resulting meshes with prior state-of-the-art methods. Second, we show the advantages of incorporating GAMesh inside the point network to directly train the network for 3D shape prediction.

\subsection{Single View Reconstruction}
\label{svr_gam_offlie}
Given a single RGB image of an object, our task is to reconstruct a 3D mesh with correct topology and accurate geometry. To this end we first train a point generation network called PSG$^{+}$ to output 2000 points. We then use meshes from implicit networks (OccNET \cite{Mescheder:2019aa} and IM-NET \cite{Chen:2018aa}) as priors for GAMesh to generate a surface for the output points.

\noindent
\textbf{Data} We use 3D models from 13 categories of ShapeNetCore.v1 \cite{shapenet2015} and renderings from 3DR2N2 \cite{Choy:2016aa} which together have been widely used for SVR \cite{Choy:2016aa,Gkioxari:2019aa,Wang:2018aa}. We sample points on the surface of the meshes for training the point network and use the same train/test split as \cite{Choy:2016aa} with renderings from all 24 viewpoints for a fair comparison.

\noindent
\textbf{Evaluation} Since our goal is to asses the quality of the reconstructed mesh, similar to \cite{Gkioxari:2019aa,Smith:2019aa}, we sample 10k points uniformly at random from the surfaces of both the predicted and the ground-truth mesh and use it to compute Chamfer distance (CD) \cite{Fan:2017aa} and F1$^{\tau}$ score \cite{Knapitsch:2017aa}. We use the same evaluation protocol as \cite{Wang:2018aa} and use a 0.57 mesh scaling factor and $\tau = 10^{-4}$ on the squared Euclidean distance. Lower is better for CD and higher is better for F1 score.

\begin{figure}[!t]
    \centering
    \captionsetup[sub]{font=footnotesize,justification=raggedright,labelformat=empty}
    \includegraphics[trim={0cm 0cm 0cm 0cm},clip,width=\linewidth, height=4.5cm]{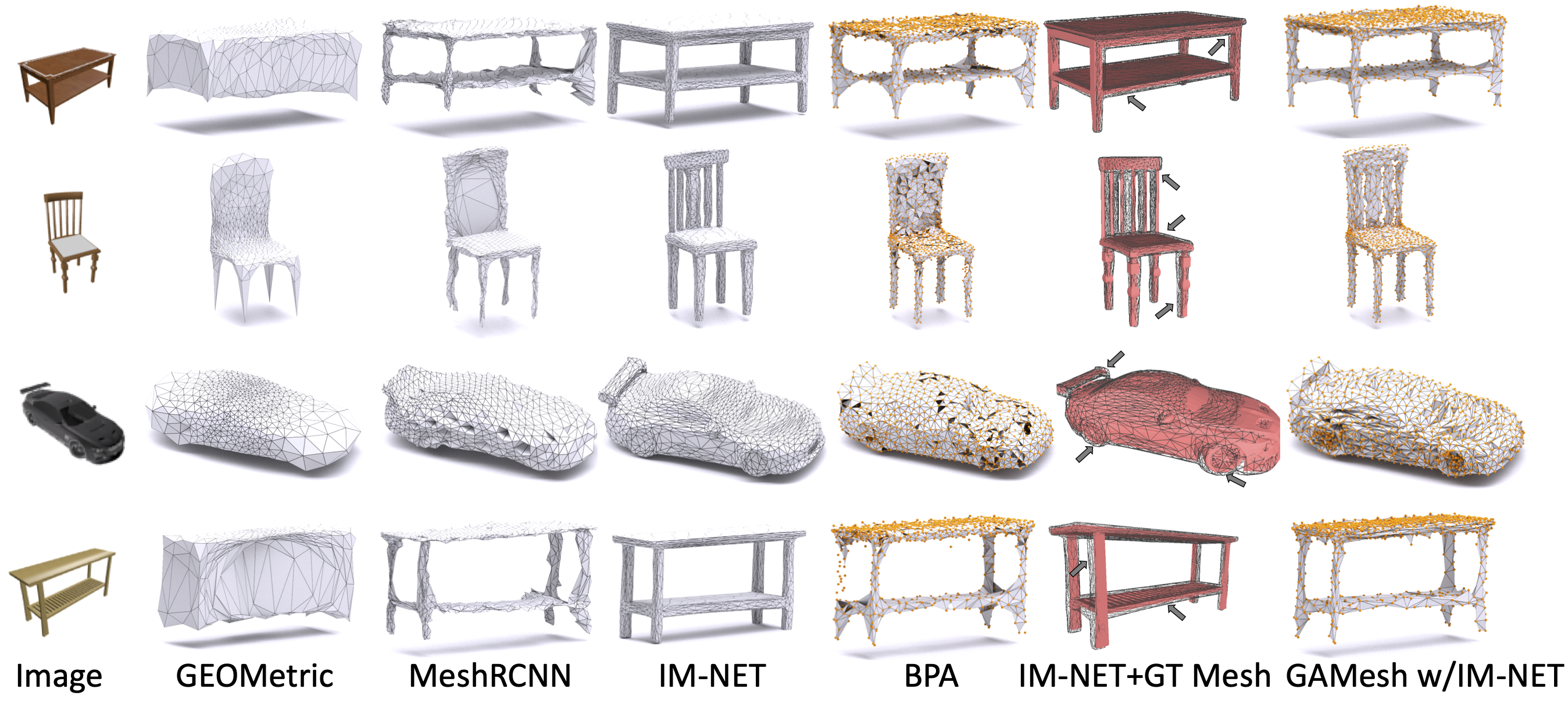}
    \caption{\textbf{Single View Reconstruction Results.} Although IM-NET reconstructs smooth meshes, it does not capture fine details such as those around the wheels of the car. Further, the output surfaces from IM-NET (wireframe) do not align well with the ground truth mesh (red) as shown by the arrows in the sixth column. GAMesh uses the output points (orange) from a point network (PSG$^{+}$) to improve the reconstructions of implicit networks. Unlike BPA, GAMesh guarantees to connect all the output points and reconstructs meshes with \emph{both} accurate geometry and correct topology. Please zoom in for details.}
    \label{fig:svr_test}
\end{figure}

\noindent
\textbf{Network \& Implementation details} We use a point network architecture where for the image encoder we use ResNet-18 \cite{He:2016aa} and for the point decoder 4 fully-connected layers of size 1024, 512, 256, (3x2000). We use ReLU non-linearity and batch normalization on the first three and tanh on the final layer. We train this network using Chamfer loss \cite{Fan:2017aa} for 30 epochs with 32 images per batch and Adam optimizer \cite{Kingma:2014aa} at a learning rate $10^{-4}$ which decays by 0.8 every 10 epochs. After training, we use GAMesh as post-processing to combine the output points from our point network (PSG$^{+}$) and the meshes from implicit networks to generate the final predicted mesh.

\noindent
\textbf{Baselines} We compare our results with several state-of-the-art SVR methods. 3DR2N2 \cite{Choy:2016aa} and MVD \cite{Smith:2018aa} are voxel based methods which reconstruct meshes with arbitrary topology. Implicit networks such as OccNET \cite{Mescheder:2019aa} and IM-NET \cite{Chen:2018aa} also generate meshes with arbitrary topology but at high resolution. PSG \cite{Fan:2017aa} outputs point predictions. P2M \cite{Wang:2018aa} and GEOMetrics \cite{Smith:2019aa} deform a template to output a 3D mesh with fixed topology (zero genus). N3MR \cite{Kato:2018aa} also deforms a template with a fixed topology but uses a differentiable renderer to train without any 3D supervision. MeshRCNN \cite{Gkioxari:2019aa} first predicts voxels with correct topology and then refines it to output a 3D mesh. We compare these methods not only on their mean CD and F1 scores on ShapeNet testset, but also on their topology and number of vertices in their output mesh. For MeshRCNN, we report the performance of their ``pretty" model as it generates topologically correct meshes. For PSG, F1 scores were computed directly using the output points \cite{Wang:2018aa}. For OccNet and IM-NET, we use the models released by authors and simplify \cite{Garland:1997aa} the meshes to approximately 2000 vertices for a fair comparison.

Apart from comparing against previous methods, we also compare our results with few ablated versions of our method. Similar to PSG, we directly evaluate the output points of our point network and report it as PSG$^{+}$. Further, instead of GAMesh, we generate surfaces for the same output points using existing meshing methods like BPA and SPR. For SPR, for the normals of the output points, we use the normal vectors of their closest points in the ground truth mesh. For both BPA and SPR, we choose the parameters which gave us the best results and reconstruct the meshes using MeshLab \cite{Cignoni:2008aa}.

\begin{table}[!b]
\centering
\caption{\textbf{Quantitative Results for SVR.} We compare several SVR methods by their output representation, $|V|$ (mean$_{\pm \text{std}}$), CD $(\text{x}10^{3})$, F1 score and topology of the output mesh. For \cite{Wang:2018aa}, $^{\dagger}$ reports the results from their paper and $^{\ddagger}$ using the model released by authors. We show that meshes reconstructed from GAMesh using IM-NET as priors achieves high fidelity and correct topology.}
\resizebox{\linewidth}{!}{
    \begin{tabular}{l|ccccc}
        \toprule
        & Representation & $|V|$ & CD ($\downarrow$)  & F1$^{\tau}(\uparrow)$ & Topology \\
        \midrule
        N3MR \cite{Kato:2018aa} & Mesh & 642$_{\pm0}$ & - & 33.80 & fixed \\
        3DR2N2 \cite{Choy:2016aa} & Voxel & - & - & 39.01 & arbitrary \\
        PSG \cite{Fan:2017aa} & Points & 1024$_{\pm0}$ & - & 48.58 & - \\
        P2M \cite{Wang:2018aa}$^{\dagger}$ & Mesh & 2466$_{\pm0}$ & - & 59.72 & fixed \\
        OccNET \cite{Mescheder:2019aa} & Implicit Field & 1998$_{\pm8}$ & 0.825 & 62.91 & arbitrary \\
        MVD \cite{Smith:2018aa} & Voxel & - & - & 66.39 & arbitrary \\
        GEOMetrics \cite{Smith:2019aa} & Mesh & 574$_{\pm99}$ & - & 67.37 & fixed \\
        IM-NET \cite{Chen:2018aa} & Implicit Field & 1999$_{\pm21}$ & 0.502 & 67.68 & arbitrary \\
        P2M \cite{Wang:2018aa}$^{\ddagger}$ & Mesh & 2466$_{\pm0}$ & 0.444 & 68.94 & fixed \\
        MeshRCNN \cite{Gkioxari:2019aa} & Mesh & 1896$_{\pm928}$ & 0.397   & 69.30 & arbitrary \\
        \midrule
        PSG$^{+}$ & Points & 2000$_{\pm0}$ & 0.424 & 56.41 & - \\
        SPR & Mesh & 9069$_{\pm929}$ & 1.206 & 59.53 & arbitrary \\
        BPA & Mesh & 1871$_{\pm27}$ & 0.399  & 70.81 & arbitrary \\
        GAMesh w/ OccNET & Mesh & 2000$_{\pm0}$ & 0.415 & 70.39 & arbitrary \\
        GAMesh w/ IM-NET & Mesh & 2000$_{\pm0}$ & 0.387 & \textbf{70.82} & arbitrary \\
        \bottomrule
    \end{tabular}
}
\label{tab:svr_summary}
\end{table}

\noindent
\textbf{Comparison with Point and Implicit Networks} Since GAMesh uses the output of both point and implicit networks, we first compare GAMesh with these two networks for single-view shape prediction. Table \ref{tab:gam_performance} shows that combining the geometry from a point network (PSG$^{+}$) and topology from implicit networks (OccNET/IM-NET) performs better than either the implicit or the point network alone. Further, we get similar F1 scores using mesh priors from either OccNet or IM-NET. This confirms that GAMesh does not require accurate mesh priors as long as they coarsely resemble the ground truth shape in terms of both topology and geometry. Please see supplementary for a detailed analysis on the effects of mesh prior on GAMesh. We also reconstruct surfaces using the ground truth meshes as prior for GAMesh. This gives us an upper bound and indicates that the performance can further be improved with an implicit network that outputs a more accurate topology.

\noindent
\textbf{Comparison with Prior Work} We now compare our results with previous works. In Table \ref{tab:svr_summary} we show that by combining the geometry from our point network (PSG$^{+}$) and topology from IM-NET we outperform previous methods both in terms of CD and F1 score using similar number of vertices. Further, when comparing GAMesh against other meshing methods, we outperform SPR by a wide margin but perform similar to BPA. BPA, however, cannot be used for surface reconstruction as it does not guarantee correct topology and often fails to connect all the points as shown in Fig. \ref{fig:svr_test} and Table \ref{tab:svr_summary}. On the contrary, GAMesh generates meshes with correct topology and guarantees to connect all the output points. Please see supplementary for more qualitative results including results on online images.

\subsection{Training Point Network with GAMesh}
\label{svr_online}

In the previous section we train the point network using Chamfer loss between the ground truth and the output \emph{points} and used GAMesh in post processing to generate meshes with uniformly distributed points. Here we show that by incorporating GAMesh inside the point network, we can reconstruct adaptive meshes with arbitrary topology. Specifically, during training, we generate a surface for the output points using GAMesh and compute a mesh loss between the ground truth and the output \emph{surface}. By backpropogating this loss through the point network, we regress on the output points such that the output surface is as close as possible to the ground truth surface. Note, GAMesh does not have any learnable parameters and is deterministic where for the same output points, it will generate the same mesh.

We demonstrate this by training two point networks to output 250 points for SVR. We train one network with Chamfer loss on the output points and the other with $\mathcal{L}_{mesh}$ (Eq. \ref{equ:mesh_loss}) on the meshes reconstructed using GAMesh. We use implicit meshes as priors for GAMesh and train both networks on renderings from two categories (chair and couch) of ShapeNet. After training we reconstruct the output meshes for both networks using GAMesh. Please see supplementary for implementation details.

\noindent
\textbf{Mesh Loss} \hspace{0.5pt} Since the point network now directly predicts a 3D surface, we need a loss function which computes an error between the predicted mesh $\mathcal{M}_2$ and the ground truth mesh $\mathcal{M}_1$. Similar to \cite{Gkioxari:2019aa,Pan:2019aa}, although we can sample points on both meshes and define a point loss (like Chamfer loss) on these resampled points, this would result in the output points being uniformly distributed thereby not fully taking advantage of the mesh representation. We on the other hand only want the predicted surface to lie close to the ground truth surface and are indifferent to the output point distribution. Hence, we define the mesh loss as:

\begin{equation}
\centering
{ \mathcal{L}_{mesh} (\mathcal{M}_1, \mathcal{M}_2) = \underset{p \in \mathcal{P}}{\sum} \Phi(p,\hat{\mathcal{Q}}) \; + \; \underset{q \in \mathcal{Q}}{\sum} \; \Phi(q,\hat{\mathcal{P})}
}
\label{equ:mesh_loss}
\end{equation}

\noindent
where $\mathcal{P}$ and $\mathcal{Q}$ are the ground truth and output points, $\mathcal{\hat{P}}$ and $\mathcal{\hat{Q}}$ are the points sampled on the ground truth mesh $\mathcal{M}_1$ and the predicted mesh $\mathcal{M}_2$ respectively, and $\Phi(a,B) = \underset{b \in B}{min} \; || a-b ||^2_2$. Essentially, $\mathcal{L}_{mesh}$ minimizes both the distance between the ground truth points to the predicted surface for \emph{high coverage} and the distance between the output points to the ground truth surface for \emph{high accuracy}. We approximate both the meshes by densely sampling 10k points on their surfaces using a differentiable mesh sampling strategy \cite{Smith:2019aa}. This replaces the expensive point-surface distance computation with the fast point-point distance computation.

\begin{figure}[!t]
    \begin{minipage}[c]{0.70\linewidth}
    \vspace{0pt}
    \centering
    \includegraphics[trim={0cm 0cm 0cm 0cm},clip,width=\linewidth,keepaspectratio]{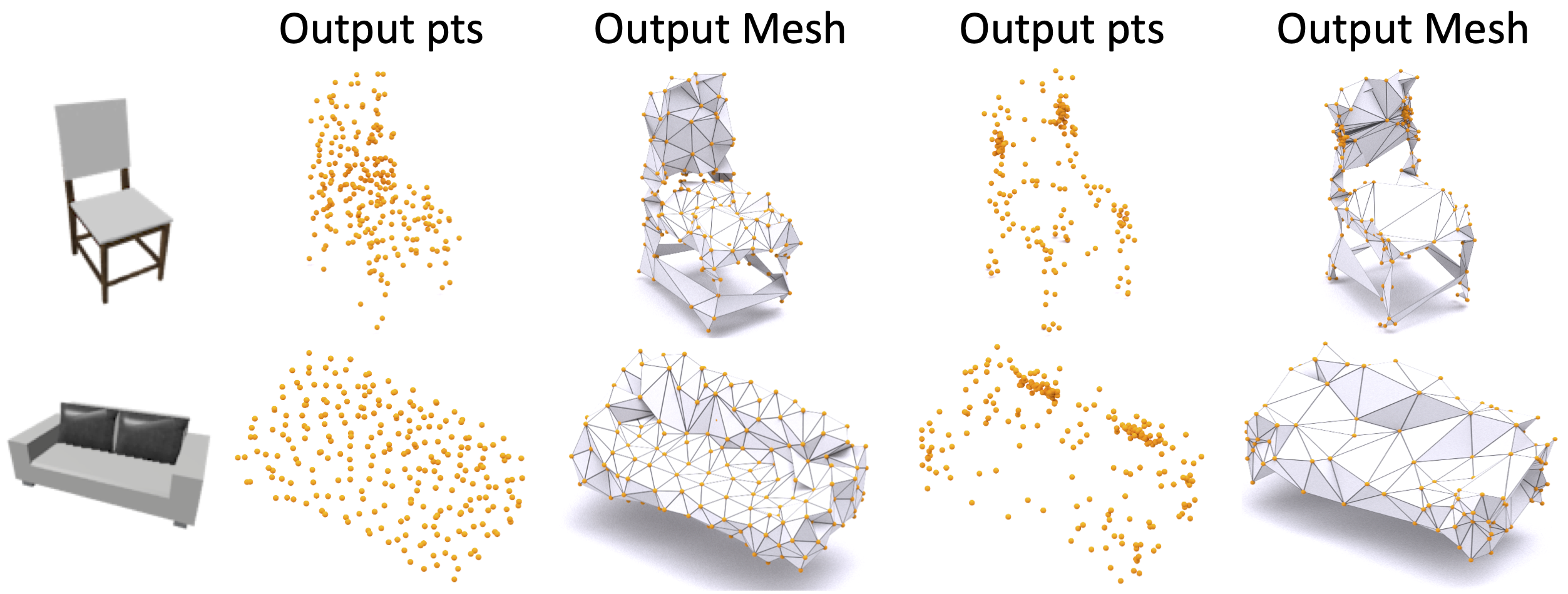}
    \scriptsize{Image \hspace{2em} Train with CD \hspace{3em} Train with Mesh Loss}
    \end{minipage}
    \hfill
    \begin{minipage}[c]{0.29\linewidth}
    \vspace{0pt}
    \centering
    \resizebox{\linewidth}{!}{
      \begin{tabular}{l|cc}
            \toprule
            \multirow{2}{*}{F1$^{\tau}$} & Train & Train \\
             & w/ CD & w/ $\mathcal{L}_{mesh}$ \\
            \midrule
            Chair & 44.64 & \textbf{48.17}\\
            Couch & 44.88 & \textbf{45.33} \\
            \midrule
            Mean & 44.76 & \textbf{46.75}\\
            \bottomrule
        \end{tabular}
    }
    \end{minipage}
\caption{\textbf{Training with GAMesh.} Training a point network using mesh loss on surfaces reconstructed by GAMesh allows us to redistribute the points towards the edges of the shape, generating adaptive meshes. On the contrary, training the same network with Chamfer loss on output points generates meshes with uniformly distributed points.}
\label{fig:GAM_online}
\end{figure}

\noindent
\textbf{Results on Test Images} We compare the output meshes from both networks in Figure \ref{fig:GAM_online}. We find training with GAMesh to perform slightly better in terms of mean F1 score. Further, while Chamfer loss uniformly distributes the points, $\mathcal{L}_{mesh}$ pushes the points towards the edges of the shape as GAMesh, being invariant to point distribution, can still reconstruct accurate surfaces. Clustering of points on the edges suggests that we could possibly reduce the number of points further, however this goes beyond the focus of this paper and is left as a potential future work.

\section{Applications}

The properties of GAMesh together with point networks allow for several other applications.

\subsection{Evaluating Point Networks}
\label{point_evaluation}
Most point based reconstruction networks \cite{Fan:2017aa,Lin:2018ab,Achlioptas:2017aa,Agarwal:2019aa} have access to ground truth meshes, but still evaluate their network by comparing the \emph{output points} with the ground truth points/surfaces. We show that evaluating the output points alone can be misleading and often wrongly implicate the network to have low performance. Evaluating the \emph{surfaces} reconstructed using the output points with the ground truth meshes (as priors for GAMesh) provides a more accurate assessment for the point network's performance.

To demonstrate this we train four point networks with the same backbone architecture but different number of output points (2500, 2000, 1000, 500) for SVR. We train these networks for the same amount of time on two categories (chair and plane) of ShapeNet. After training we not only report the CD on the output points, but also evaluate the surfaces (using F1 score) reconstructed using BPA, SPR and GAMesh using the ground truth meshes as priors. We also measure the error from networks with 2500 and 500 output points and report it as $\Delta$ in Table \ref{tab:pointEvaluation}.

\begin{table}[!b]
\centering
\caption{\textbf{Evaluating Point Networks for SVR.} We compare four point networks with different number of output points by CD $(\text{x}10^{3})$ on the output points, F1 scores on the surfaces reconstructed using BPA, SPR \& GAMesh and the mean $\%$ of unreferenced vertices in these reconstructed surfaces. Analyzing the network's performance through the output points alone can often be misleading. Evaluating the surfaces reconstructed by GAMesh using the output points gives a more accurate assessment.}
\resizebox{\linewidth}{!}{
  \begin{tabular}{ll|cccc|c}
    \toprule
    & & \multicolumn{4}{c|}{$|V|$} &  \multirow{2}{*}{$\Delta (\downarrow)$} \\
    \cmidrule(lr){3-6}
    & & 2500 & 2000 & 1000 & 500 & \\
    \midrule
    \multirow{4}{*}{\rotatebox[origin=c]{90}{Chair}}
    & CD ($\downarrow$) & 4.60 & 5.01 & 6.76 & 9.13 & 98.47$\%$\\
    & F1$^{\tau}$ (BPA) ($\uparrow$) & 68.77$_{5.9\%}$ & 69.08$_{5.8\%}$ & 65.63$_{5.4\%}$ & 57.39$_{5.4\%}$ & 19.82$\%$\\
    & F1$^{\tau}$ (SPR) ($\uparrow$) & 56.64$_{99.9\%}$ & 55.30$_{99.9\%}$ & 50.39$_{100\%}$ & 44.64$_{99.9\%}$ & 26.88$\%$ \\
    & F1$^{\tau}$ (GAMesh) ($\uparrow$) & \textbf{72.74$_{0\%}$} & \textbf{73.32$_{0\%}$} & \textbf{70.95$_{0\%}$} & \textbf{69.25$_{0\%}$} & \textbf{4.79$\%$} \\
    \midrule
    \midrule
    \multirow{4}{*}{\rotatebox[origin=c]{90}{Plane}}
    & CD ($\downarrow$) & 2.23 & 2.36 & 3.26 & 4.61 & 106.7$\%$\\
    & F1$^{\tau}$ (BPA) ($\uparrow$) & 86.07$_{6.2\%}$ & 85.86$_{6.2\%}$ & 84.84$_{5.1\%}$ & 83.43$_{4.0\%}$ & 3.06$\%$\\
    & F1$^{\tau}$ (SPR) ($\uparrow$) & 68.92$_{99.9\%}$ & 65.82$_{99.9\%}$ & 61.77$_{99.9\%}$ & 53.33$_{100\%}$ & 22.62$\%$\\
    & F1$^{\tau}$ (GAMesh) ($\uparrow$) & \textbf{88.59$_{0\%}$} & \textbf{88.78$_{0\%}$} & \textbf{87.89$_{0\%}$} & \textbf{86.25$_{0\%}$} & \textbf{2.64$\%$}\\
    \bottomrule
  \end{tabular}
}
\label{tab:pointEvaluation}
\end{table}

\noindent
\textbf{Results} We draw the following conclusions from Table \ref{tab:pointEvaluation}. (a) Networks trained using low number of output points have higher CD ($\Delta${\raise.17ex\hbox{$\scriptstyle\mathtt{\approx}$}}100\%) incorrectly suggesting that those networks are weak and their output points are not reconstructed properly. However, analyzing their F1 scores suggests that all surfaces are similar (small $\Delta$) and all networks perform equally well. This shows that we cannot solely evaluate the output points to assess the network's performance and should also evaluate the meshes reconstructed using those output points. (b) Meshes reconstructed using BPA and SPR on average fail to include 5.6$\%$ and 99$\%$ of the output points respectively. Hence, they should not be used to mesh the points from a point network. Both BPA and SPR are highly sensitive to the input point distribution and require dense and uniform points for an accurate reconstruction. (c) For all the four point networks, meshes reconstructed using GAMesh have a higher F1 score than BPA and SPR. This suggests that the points from all networks capture accurate geometry but the meshing algorithms (BPA and SPR) reconstructed poor surfaces. While BPA cannot guarantee correct topology as it fails to connect all output points, SPR cannot guarantee correct geometry as it interpolates the output points using additional points (Fig. \ref{fig:GAM_vs_bpa/spr}). GAMesh is independent of both point density \& distribution and always includes only and all the output points of the point network in the reconstructed mesh. It guarantees correct topology \& geometry and hence can be used in post-processing to decouple the reconstruction error from the network error. Please see the supplementary for tests on the robustness of GAMesh.

\begin{figure}[!t]
    \centering
    \captionsetup[sub]{font=scriptsize,justification=raggedright}
    \begin{subfigure}[!t]{0.24\linewidth}
        \centering
        \includegraphics[trim={6cm 4cm 6cm 4cm},clip,width=\linewidth, keepaspectratio]{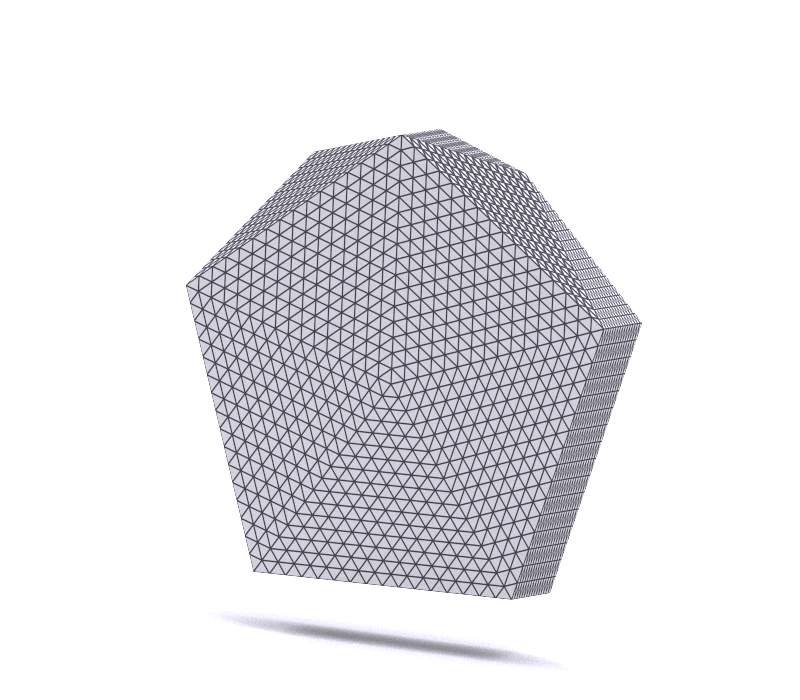}
        \caption{Original Mesh}
    \end{subfigure}
    \begin{subfigure}[!t]{0.24\linewidth}
        \centering
        \includegraphics[trim={6cm 4cm 6cm 4cm},clip,width=\linewidth, keepaspectratio]{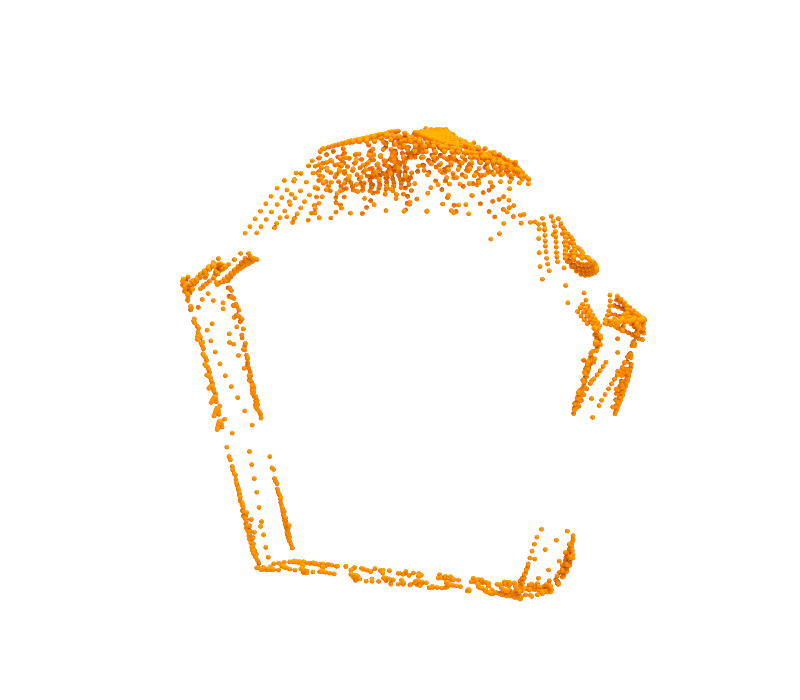}
        \caption{Output from \cite{Agarwal:2019aa}}
    \end{subfigure}
    \begin{subfigure}[!t]{0.24\linewidth}
        \centering
        \includegraphics[trim={6cm 4cm 6cm 4cm},clip,width=\linewidth, keepaspectratio]{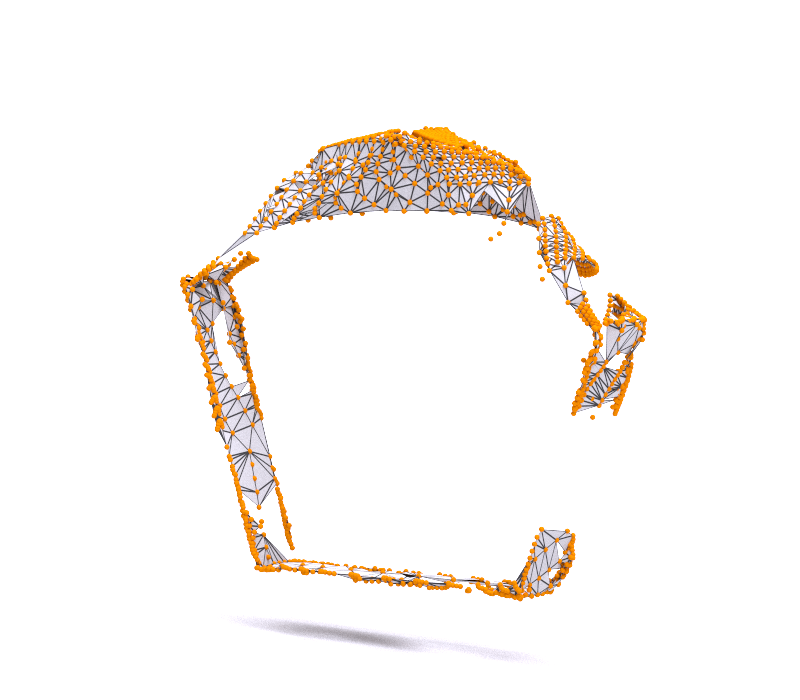}
         \caption{BPA}
    \end{subfigure}
    \begin{subfigure}[!t]{0.24\linewidth}
        \centering
        \includegraphics[trim={6cm 4cm 6cm 4cm},clip,width=\linewidth, keepaspectratio]{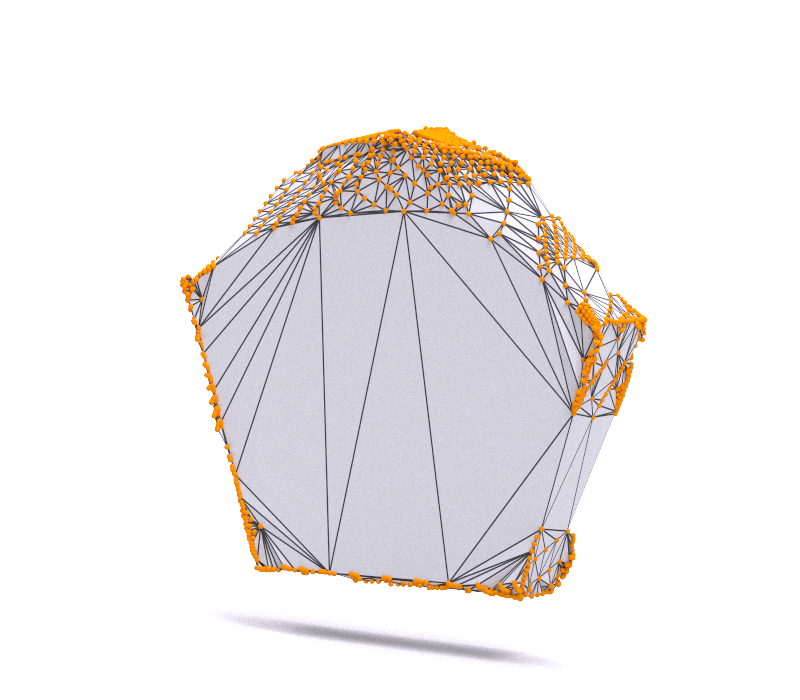}
        \caption{GAMesh}
    \end{subfigure}
    \caption{\textbf{Meshing Sparse Point Clouds.} GAMesh can be used to generate surfaces for point networks which output sparse point clouds.}
    \label{fig:sparse_points}
\end{figure}

\subsection{Reconstructing Surfaces for Sparse Points}
As GAMesh is indifferent to both point density \& distribution, it can be used with various point networks which output sparse point clouds \cite{Wang:2020ab,Loizou:2020aa,Agarwal:2019aa}. As opposed to BPA or SPR, GAMesh guarantees to reconstruct an accurate surface connecting all the output points using the ground truth meshes as priors. Such meshes can then be used for qualitative and/or quantitative analysis of the network (Fig. \ref{fig:sparse_points}).

%% file: tex_files/conclusion.tex
\section{Limitation and Conclusion}

In this paper, we introduce GAMesh, a new meshing algorithm to generate a surface for the output points of a point network using a mesh prior. GAMesh decouples geometry from topology by making the point network solely responsible for geometry and the mesh prior responsible for topology. We show the benefits of such a disentanglement for single-view shape prediction and fair evaluation of point networks. Further, unlike traditional surface reconstruction algorithms, GAMesh is independent of the density and distribution of the output points and guarantees to reconstruct a surface with correct topology and geometry.

As GAMesh aims to preserve the geometry from point networks, the resulting meshes are often less smooth than implicit methods. Using GAMesh to generate adaptive but smooth meshes could be interesting future direction. GAMesh also requires a mesh prior which is aligned and coarsely resembles the ground truth shape. Although this may seem as a strong assumption, in our experience such mesh priors, if not already present, can be obtained from other reconstruction methods. Therefore, we believe GAMesh is an attractive meshing algorithm for deep point networks.